\documentclass[10pt,conference]{IEEEtran}
\IEEEoverridecommandlockouts % show footnote comments
\usepackage{blkarray}                                      % to support matrix
\usepackage{algpseudocode}                                 % new algorithm package
\usepackage{algorithm}
\usepackage{graphicx}                                      % include pdf figures
\usepackage{amsmath}
\usepackage{amssymb}
\usepackage{amsfonts}
\usepackage{amsthm}
\usepackage[mathcal]{eucal}
\usepackage{mathrsfs}
\usepackage{booktabs}
\usepackage{enumerate}
\usepackage{multirow}
\usepackage[subrefformat=parens,farskip=0pt,justification=centering]{subfig}
\usepackage{color}
\usepackage{cite}                                          % more citations in one bracket
\usepackage{comment}                                       % use comment
\usepackage{soul}                                          % use highlight command \hl{}
\soulregister\cite7
\soulregister\ref7
\soulregister\pageref7
\usepackage{etoolbox}                                      % commands \newtoggle, \toggletrue, \iftoggle
\usepackage{url}
\usepackage{nth}                                           % nth command
\usepackage{bm}                                            % bm command
\usepackage{courier}
\usepackage{makecell}
\usepackage{balance}
\usepackage{threeparttable}
\usepackage[bookmarks=false]{hyperref}
\hypersetup{
    colorlinks = true,
    citecolor  = blue,
    linkcolor  = blue,
    urlcolor   = blue,
}
\usepackage{tikz}
\usetikzlibrary{positioning,calc,fit,decorations.pathmorphing,shapes.geometric, shapes.gates.logic.US, calc}
\usetikzlibrary{arrows,arrows.meta,decorations.markings,shapes,shapes.arrows}
\usetikzlibrary{backgrounds}
\usepackage{filecontents}                                  % support to pgfplots
\usepackage{pgfplots}
\usepackage{pgfplotstable}
\usepackage{scalefnt}
\usepgfplotslibrary{groupplots}
\pgfplotsset{compat=newest}
\usepackage{caption}
\usepackage{cleveref}
\Crefformat{figure}{Fig.~#2#1#3}                           % "Fig.", instead of "Figure"
\Crefname{subfigure}{Fig.}{Figs.}
\Crefname{figure}{Fig.}{Figs.}
\Crefformat{table}{TABLE~#2#1#3}                           % "TABLE", instead of "Table"
\definecolor{CUpurple}{RGB}{117,15,109}
\definecolor{CUgold}{RGB}{221,163,0}
\definecolor{CUribbon}{RGB}{244,223,176}
\definecolor{CUblack}{RGB}{34,24,21}
\definecolor{USTgold}{RGB}{153,102,0}
\definecolor{USTyellow}{RGB}{204,153,0}
\definecolor{USTyellowlight}{RGB}{255,212,0}
\definecolor{USTorange}{RGB}{255,166,26}
\definecolor{USTblue}{RGB}{0,51,102}
\definecolor{USTmiddle}{RGB}{0,116,188}
\definecolor{USTlight}{RGB}{99,202,225}
\definecolor{USTgray}{RGB}{204,204,204}
\definecolor{USTred}{RGB}{237,27,47}
\definecolor{USTdarkred}{RGB}{124,35,72}

\usepackage{tcolorbox}
\tcbuselibrary{skins,breakable}
    {\endtcolorbox}

\usepackage[top=0.75in,bottom=0.75in,left=0.55in,right=0.55in]{geometry}
\newcommand{\subparagraph}{}
\usepackage{titlesec}
\titlespacing*{\section}{0pt}{1.8ex plus .2ex minus .2ex}{0.4ex plus .2ex}
\titlespacing*{\subsection}{0pt}{1.0ex plus .2ex minus .2ex}{0.2ex plus .2ex}

\crefname{mytheorem}{Theorem}{Theorems}
\crefname{mylemma}{Lemma}{Lemmas}
\crefname{myclaim}{Claim}{Claims}
\crefname{myproperty}{Property}{Properties}
\crefname{mycorollary}{Corollary}{Corollaries}

\algrenewcommand\textproc{\texttt}

\makeatletter
\let\OldStatex\Statex
\renewcommand{\Statex}[1][3]{%
  \setlength\@tempdima{\algorithmicindent}%
  \OldStatex\hskip\dimexpr#1\@tempdima\relax
}
\makeatother

\RequirePackage[normalem]{ulem} %DIF PREAMBLE
\RequirePackage{color}\definecolor{RED}{rgb}{1,0,0}\definecolor{BLUE}{rgb}{0,0,1} %DIF PREAMBLE
\newcommand{\etal}{\textit{et al}.}

\begin{document}
\title{PICBench: Benchmarking LLMs for Photonic Integrated Circuits Design}

\iftrue 
\author{
\IEEEauthorblockN{Yuchao Wu}
\IEEEauthorblockA{HKUST(GZ)}
\and
\IEEEauthorblockN{Xiaofei Yu}
\IEEEauthorblockA{HKUST(GZ)}
\and
\IEEEauthorblockN{Hao Chen}
\IEEEauthorblockA{HKUST(GZ)}
\and
\IEEEauthorblockN{Yang Luo}
\IEEEauthorblockA{HKUST(GZ)}
\and
\IEEEauthorblockN{Yeyu Tong}
\IEEEauthorblockA{HKUST(GZ)}
\and
\IEEEauthorblockN{Yuzhe Ma}
\IEEEauthorblockA{HKUST(GZ)}
}
\fi

\maketitle

\begin{abstract}
While large language models (LLMs) have shown remarkable potential in automating various tasks in digital chip design, the field of Photonic Integrated Circuits (PICs)—a promising solution to advanced chip designs—remains relatively unexplored in this context. The design of PICs is time-consuming and prone to errors due to the extensive and repetitive nature of code involved in photonic chip design.
In this paper, we introduce \textbf{PICBench}, the first benchmarking and evaluation framework specifically designed to automate PIC design generation using LLMs, where the generated output takes the form of a netlist.
Our benchmark consists of dozens of meticulously crafted PIC design problems, spanning from fundamental device designs to more complex circuit-level designs. 
It automatically evaluates both the syntax and functionality of generated PIC designs by comparing simulation outputs with expert-written solutions, leveraging an open-source simulator.
We evaluate a range of existing LLMs, while also conducting comparative tests on various prompt engineering techniques to enhance LLM performance in automated PIC design.
The results reveal the challenges and potential of LLMs in the PIC design domain, offering insights into the key areas that require further research and development to optimize automation in this field.
Our benchmark and evaluation code is available at https://github.com/PICDA/PICBench.
\end{abstract}

\pagestyle{empty}

\section{Introduction}
\label{sec:intro}
Photonic Integrated Circuits (PICs) represent a groundbreaking advancement in chip design, harnessing the properties of light to enable faster data processing and greater energy efficiency. 
As the demand for high-performance computing and communication systems continues to rise, PICs have emerged as a critical solution to meet these requirements \cite{sun2015single,kitayama2019novel,lu2024empowering,xu2024large}. 
However, unlike the maturity of electronic design automation (EDA) tools, the development of photonic design automation (PDA) tools capable of supporting automated design pipelines for circuit simulation and layout remains at an early stage.
The design and layout of photonic circuits and components still heavily rely on manual input, which introduces significant inefficiencies. 
Photonic designs are inherently complex, often requiring repetitive, low-level coding for devices and connections. 
This process is time-intensive and prone to human error, particularly as the size and complexity of the designs increase.
Consequently, there is an urgent need for a comprehensive set of tools to fully automate photonic circuit design and layout processes. 
Advancements in large language models (LLMs) \cite{team2023gemini,TheC3,achiam2023gpt} offer a promising opportunity to address these challenges and accelerate the development of PDA solutions.

Recently, LLMs have demonstrated significant potential in automating code generation for hardware designs, which offers substantial support to engineers in designing and verifying these systems. 
RTLLM \cite{lu2024rtllm} introduced a benchmark framework comprising 30 designs spanning diverse complexities and scales for Verilog generation. 
Then VerilogEval \cite{liu2023verilogeval} proposed an extensive dataset of 156 problems and a robust testing procedure to facilitate the systematic evaluation of generated code. 
Beyond Verilog generation, SPICEPilot \cite{vungarala2024spicepilot} investigated the capabilities of LLMs in generating SPICE code. 
Another work ChatEDA \cite{wu2024chateda} demonstrated the ability to generate code for interacting with EDA tools using natural language instructions. 

Nevertheless, the application of LLM in photonic circuit design has been limited to a few works. 
Li \etal \cite{li2023english} utilized LLM generating FDTD code for simulating the photonic crystal surface emitting laser (PCSEL) structure and AI code for subsequent optimizations of the PCSEL model. 
However, their approach was not fully automated, as it relied heavily on human experts to iteratively specify requirements and debug errors.
Liu \etal \cite{liu2024towards} presented an automated framework that translates natural language prompts into Python code capable of generating GDSII files using an open-source library. 
However, their framework was tested on only seven simple device designs, leaving the performance and scalability of LLM-based solutions insufficiently evaluated. 
The absence of a reliable and automated testing framework, coupled with limited datasets and the lack of a standardized benchmark, significantly hinders both the development and fair evaluation of LLM solutions in PICs design. 
To address these challenges, a comprehensive benchmark is needed—one that encompasses a wide range of design problems, includes a reliable and automated evaluation framework to minimize testing variance, and clearly distinguishes the correctness and efficiency of solutions.

\begin{figure}[!tb]
    \centering
    \subfloat{%
      \includegraphics[width=0.44\textwidth]{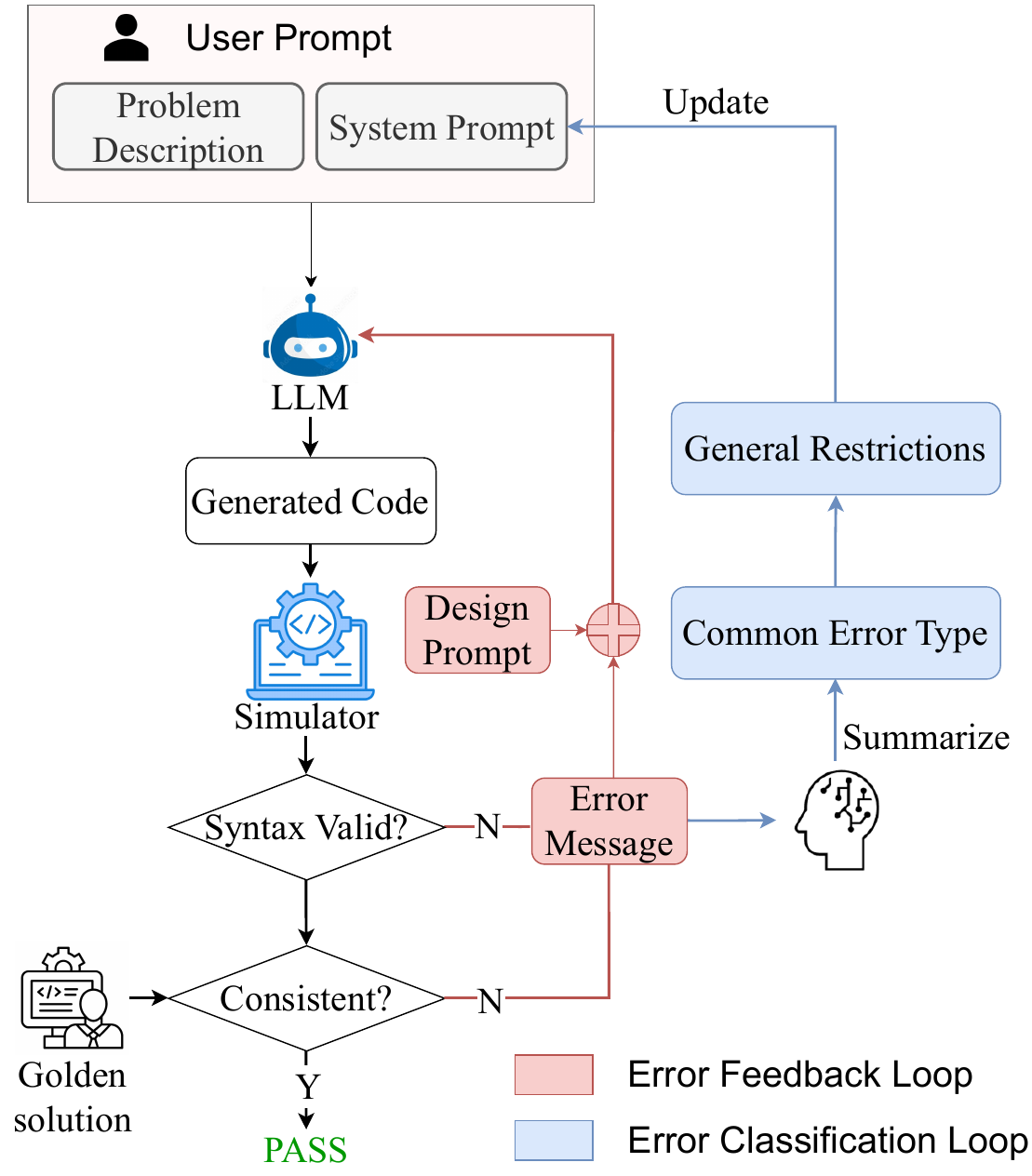}
      }
      \caption{PICBench framework that automated design generation and evaluation.}
      \label{fig:flow}
\end{figure}

% \begin{figure*}[!tb]
%     \centering
%     \subfloat{%
%       \includegraphics[width=0.98\textwidth]{fig/framework.pdf}
%       }
%       \caption{PICBench framework that automated design generation and evaluation. Users only need to provide their problem descriptions.}
%       \label{fig:flow}
% \end{figure*}

In this paper, we introduce PICBench, an open-source and comprehensive benchmark for PIC design using natural language to generate simulation-ready netlists.
The benchmark includes 24 meticulously crafted PIC design problems, covering a wide range of design complexities and scales. 
Each problem features clear descriptions and is accompanied by ground-truth designs created by human experts, serving as a golden result for evaluation.
Leveraging an open-source simulator SAX \cite{sax2023}, PICBench enables efficient and automated evaluation of any LLM-generated results.

Our contributions are summarized as follows:
\begin{itemize}
    \item We introduce PICBench, the first comprehensive open-source benchmark for PIC design using LLMs, comprising 24 carefully designed PIC design problems.
    \item We proposed a simple but efficient feedback-based method that further enhances the model’s proficiency in PIC design tasks.
    \item We exhaustively evaluated the state-of-the-art commercial LLMs with our benchmark on both syntax and functionality.
\end{itemize}

\section{Preliminary}
\label{sec:prelim}

In this section, we will first introduce the open-source PIC simulator SAX \cite{sax2023} and then present our description of the PIC design task based on natural language instructions.

\subsection{SAX}
SAX is a Python library designed for S-parameter-based circuit simulation and optimization in the frequency domain, leveraging JAX for automatic differentiation and GPU acceleration. 
It provides a functional approach to modeling photonic integrated circuits, allowing users to define components and circuits using standard Python functions and dictionaries.
Given a JSON netlist specifying input/output ports, required components, their configurations, and detailed interconnections, SAX can efficiently perform mathematical analysis and simulate circuit behavior.
\subsection{Task Description}
Here we describe the PIC design task based on natural language instructions as follows:
\begin{itemize}
\item[$\bullet$]
Given the natural language description of the desired circuit functionality and specified configurations, the objective is to understand and respond to user requirements and generate the simulate-ready netlist of this design.
\end{itemize}
\section{PICBench}
\label{sec:method}
In this section, we will first present an overview of our PICBench framework and then introduce the details.

\subsection{Framework}
\Cref{fig:flow} illustrates PICBench's flow: user provides a natural language description of their PIC design task to the LLM. 
The LLM’s output, a simulation-ready netlist, is then directly fed into SAX for simulation.
If the simulation tool reports errors, the tool’s outputs are returned to the LLM as a new prompt with a request to rectify the errors. Simultaneously, the error information is reviewed and summarized into restrictions by human inspection, which are then incorporated into the initial system prompt. 
If the simulation tool successfully generates a frequency response, PICBench tests the functionality of the generated design by comparing it with the golden results. 
The process terminates when both syntax and functionality tests are passed.
Otherwise, the process iterates up to a user-specified number of trials.

% \begin{figure*}[!tb]
%     \centering
%     \subfloat{%
%       \includegraphics[width=0.95\textwidth]{fig/framwork.pdf}
%       }
%       \caption{PICBench framework that automated design generation and evaluation. Users only need to provide their problem descriptions.}
%       \label{fig:flow}
% \end{figure*}

\subsection{Detail Description of Our Benchmark}
PICBench collects 24 meticulously crafted, commonly encountered PIC design problems, spanning a wide range of design complexities and scales. 
\Cref{tab:tb_des} shows the detailed description of all 24 problems provided in our benchmark, including 6 optical computing circuits, 7 optical interconnects circuits, 9 optical switches, and 2 fundamental devices.
These problems exhibit significant variation in their target functionalities, including examples such as optical switches, optical modulators, and more.
In addition to the diverse functionalities, our benchmark also exhibits rich variation in the implementation. 
For instance, optical switches in our benchmark share the common functionality of switching signal connections. 
However, we include all widely used architectures of optical switches, such as crossbar, Spanke, Benes, and Spanke-Benes, with configurations ranging from $4 \times 4$ to $8 \times 8$.
Notably, we do not include any purely device-level design problems in our collection, as these lack connections, and the components section in each netlist inherently addresses device-level designs.
The foundational devices we include are not simple device components. 
They involve connections among more than two components and can serve as the basis for constructing more complex circuits.

\begin{table*}[htb!]
    \centering
    \renewcommand{\arraystretch}{1.2}
    \caption{Benchmark Description}
    \label{tab:tb_des}
    \resizebox{\textwidth}{!}{ 
    \begin{tabular}{llcc}
        \toprule
        \multicolumn{2}{c}{\textbf{Design}}& \textbf{Description} \\
        \midrule
        \multirow{7}{*}{\textbf{Optical Computing}} 
         & Clements\_$4 \times 4$ & A $4 \times 4$ MZI mesh arranged using the Clements method \\
         & Clements\_$8 \times 8$ & An $8 \times 8$ MZI mesh arranged using the Clements method \\
         & Reck\_$4 \times 4$ & A $4 \times 4$ MZI mesh arranged using the Reck method \\
         & Reck\_$8 \times 8$ & An $8 \times 8$ MZI mesh arranged using the Reck method \\
         & NLS & A Non-Linear Sign gate with a signal channel and two additional ancilla channels \\
         & U-matrix block & A fundamental block representing a $2 \times 2$ unitary matrix of arbitrary values \\
         \midrule
         \multirow{7}{*}{\textbf{Optical Interconnects}} \
         & Direct modulator& An optical direct modulator  \\
         & QPSK modulator & An optical QPSK modulator  \\
         & 8-QAM modulator& An optical 8-QAM modulator \\
         & 64-QAM modulator& An optical 64-QAM modulator  \\
         & WDM\_mux & A WDM multiplexer \\
         & WDM\_demux & A WDM demultiplexer \\
         & Optical hybrid & A 90$^{\circ}$ optical hybrid\\
         \midrule
         \multirow{9}{*}{\textbf{Optical Switch}} 
         & OS\_$2 \times 2$ & A fundamental $2 \times 2$ optical switch\\
         & Crossbar\_$4 \times 4$ & A $4 \times 4$ optical switching network based on Crossbar architecture \\
         & Crossbar\_$8 \times 8$ &  An $8 \times 8$ optical switching network based on Crossbar architecture  \\
         & Spanke\_$4 \times 4$ &  A $4 \times 4$ optical switching network based on Spanke architecture \\
         & Spanke\_$8 \times 8$ & An $8 \times 8$ optical switching network based on Spanke architecture \\
         & Benes\_$4 \times 4$ & A $4 \times 4$ optical switching network based on Benes architecture \\
         & Benes\_$8 \times 8$  & An $8 \times 8$ optical switching network based on Benes architecture  \\
         & Spanke–Benes\_$4 \times 4$  & A $4 \times 4$ optical switching network based on Spanke–Benes architecture  \\
         & Spanke–Benes\_$8 \times 8$ &  An $8 \times 8$ optical switching network based on Spanke–Benes architecture \\
         \midrule
         \multirow{2}{*}{\textbf{Fundamental Devices}} 
         & MZM & A Mach-Zehnder modulator\\
         & MZI\_ps & A Mach-Zehnder interferometer with a phase shifter \\
        \bottomrule
    \end{tabular}
    }
\end{table*}

\begin{figure}[!tb]
    \centering
    \subfloat{%
      \includegraphics[width=0.48\textwidth]{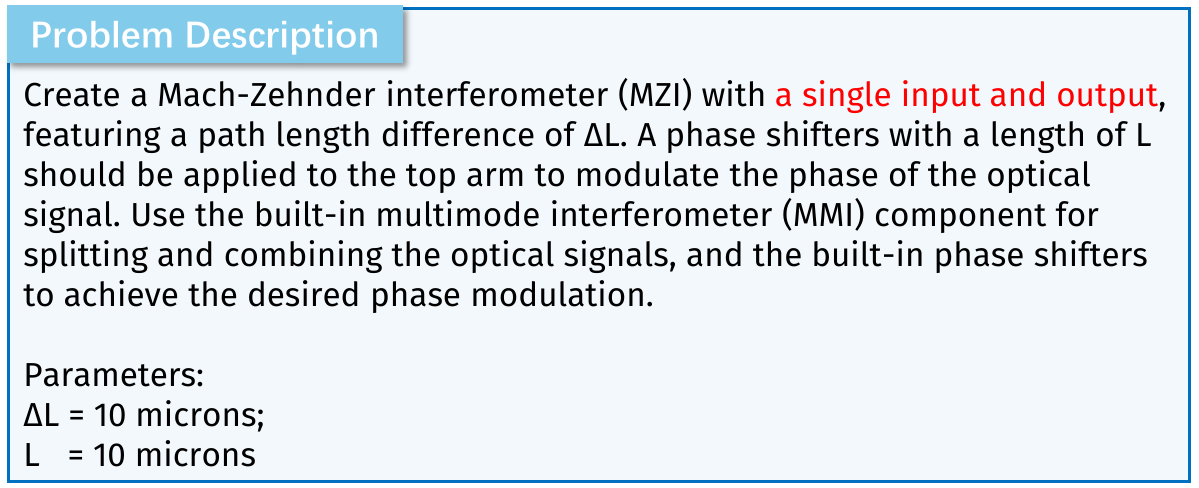}
      }
      \caption{Example of problem description.}
      \label{fig:problem_des}
\end{figure}

For each design, we provide three key components: a detailed problem description, the correct design, and its corresponding frequency response.
The problem description is a natural language description of the desired circuit functionality, including the required configurations and the number of input and output ports, as illustrated in \Cref{fig:problem_des}, which provides an example of the MZI\_ps problem.
% To ensure the correctness and reliability of the problem description, human designer is tasked with generating corresponding designs based on these descriptions. 
% If the description is unambiguous and allows for the correct design to be accurately created, it is considered valid and the manually created correct design is regarded as the golden solution for evaluation.
Human designers then manually craft the correct design based on the description, producing golden solution for evaluation.
To streamline the evaluation process, the correct design is subsequently fed into the simulator, and its frequency response is directly saved. 
This frequency response serves as a reference for verifying the correctness of the design’s functionality.

\subsection{Code Generation and Evaluation}
\label{sec:code_gen_eval}
\begin{figure}[!tb]
    \centering
    \subfloat{%
      \includegraphics[width=0.48\textwidth]{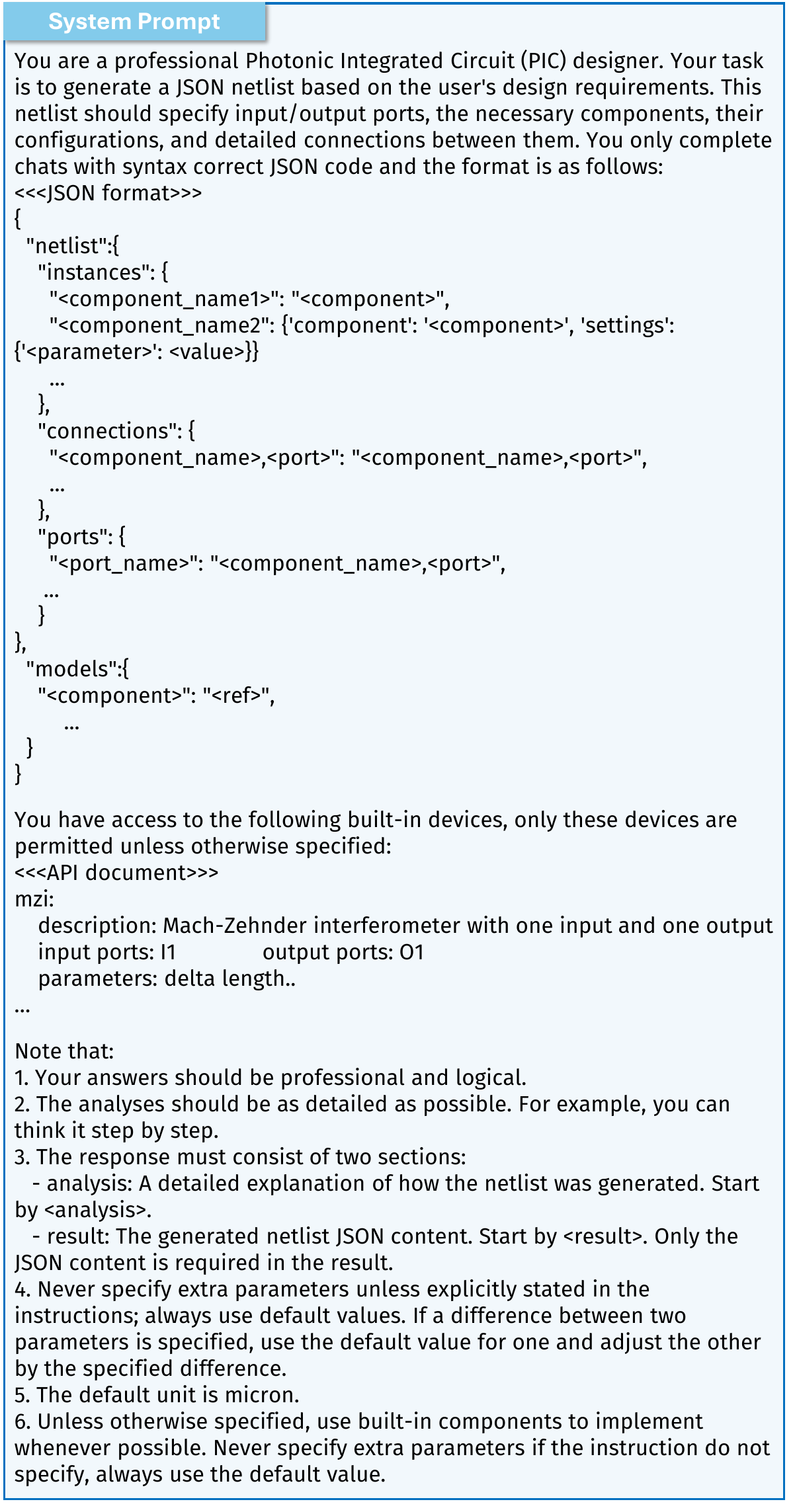}
      }
      \caption{System prompt template for code generation.}
      \label{fig:sys_prompt}
\end{figure}

Since the netlist required by SAX does not follow a general format, we designed a system prompt template to maximize the efficiency of instructing LLMs to generate high-quality, error-free designs. 
As shown in \Cref{fig:sys_prompt}, the template consists of three components: 
\begin{enumerate}
    \item {\textbf{Required format}:} This part provides the schema for the required format. It defines the structure of the netlist to ensure compliance with SAX's requirements.
    \item {\textbf{API document}:} This part includes detailed documentation of the built-in components provided by SAX or defined by us. It specifies port definitions and configurable parameters, offering a comprehensive reference for implementing various components within the netlist.
    \item {\textbf{Restrictions}:} This part establishes a set of clear restrictions designed to standardize and streamline the generation of netlist JSON content.
\end{enumerate}

% By adhering to these guidelines, the generation process ensures consistent and logical outputs, maintaining high levels of professionalism and clarity. 
This structured approach promotes uniformity across all generated netlists, minimizes errors and ambiguity, and ensures that the outputs are both precise and efficient.

Once the design is successfully generated, its evaluation is automatically conducted, focusing primarily on two aspects: syntax and functionality.

Syntax correctness is the most fundamental requirement for ensuring logical functionality and executable code, as functionality cannot be assessed without first confirming syntactic validity. 
To verify syntax, the design is tested using the SAX simulator. 
If no errors are detected and a frequency response is successfully generated, the syntax is considered valid.

Functionality correctness is also evaluated through simulation to determine whether the generated design performs as expected.
However, unlike traditional testbenches used in RTL design \cite{lu2024rtllm}, where specific input signals are crafted to verify the correctness of outputs, our simulations are conducted in the frequency domain.
In this context, the input corresponds to a range of frequencies rather than discrete signals, and the response at any single wavelength alone does not provide a conclusive indication of success or failure, nor does it enable precise and efficient feedback.
% and the response at individual frequencies lacks standalone significance.
Therefore, we simply compare the simulation results between generated code completions and golden reference solutions.
Since the built-in components are limited, we manually construct all required components based on the descriptions provided in the API document, ensuring that all problems in our benchmark are successfully evaluated.

\subsection{Error Classification Loop}
\label{sec:Error_class}
The PIC netlist design generation problem involves a specialized language and task that is rarely encountered during the pre-training of existing LLMs, leading to inherent limitations in their performance on such tasks. 
Despite employing in-context learning and providing several examples, LLMs often confuse the PIC netlist format with other netlist formats. 
For instance, in a PIC netlist, each port can only be connected once, and duplicate connections to the same port are prohibited. 
However, LLMs frequently generate connections containing multi-pin nets, as seen in traditional VLSI netlists, which is incorrect.

To effectively employ LLMs in PIC design generation, we employ an automatic error classification feedback method. 
Since the specific aspects causing LLM failures are uncertain, errors are iteratively inspected and summarized during the generation process. 
For each conversation, if a syntax error is detected, a human expert inspects the error information, identifies common errors, and summarizes them into general restrictions to prevent the recurrence of similar errors.
\Cref{tab:class_restrictions} summarize all the common error types we collected during our trials and the corresponding restrictions that we collect as prompt.
These domain-specified restrictions are then integrated into system prompt to improve understanding of PIC modules and coding styles, and provide valuable insights into the primary reasons for failures.
This prompt-tuning approach effectively addresses poor code generation, significantly enhancing LLM performance for PIC design tasks.

\begin{table}[!t]
\centering
\caption{Restrictions for the PIC design task, listing the main failure types and corresponding constraints to maintain valid syntax.}
\label{tab:class_restrictions}
\begin{tabular}{|p{3cm}|p{5.5cm}|}
\hline
\textbf{Failure Types} & \textbf{Restrictions}\\
\hline
Use undefined models &
Only built-in devices are permitted
unless otherwise specified; never
use undefined models. \\
\hline
Bind the I/O ports &
Input or output ports in the
\texttt{ports} section represent only the
system’s start or end points; they must
not appear in any internal connections. \\
\hline
Mess up ‘Instances’
and ‘models’ part &
When specifying built-in components,
the model reference must appear in
the \texttt{models} section like
\texttt{``...\,: "<ref>"''} rather
than \texttt{``"<ref>" : ...''}.
The \texttt{instances} section
only instantiates these components. \\
\hline
Extra contents found
in JSON &
Only the required JSON netlist elements
should appear in the output.
Do not include comments, advice,
or code block markings. \\
\hline
Duplicate connections
to the same port &
Each port can only be connected once;
duplicate connections to the same port
are prohibited. \\
\hline
Wrong connections for
dangling ports &
If a specific port mapping is not
explicitly required, omit it rather
than introducing arbitrary or unused
port names. \\
\hline
Wrong ports number &
The total number of input and output ports
must align with the design specification.
Each input port typically starts with \texttt{I},
and each output port with \texttt{O}. \\
\hline
Wrong ports &
Ensure all \texttt{connections} and
\texttt{ports} are valid and consistent
with the defined \texttt{instances} and
\texttt{models}. Do not generate invalid
or undefined mappings. \\
\hline
Wrong component name &
Underscores are prohibited
in component names. \\
\hline
Other syntax error & $\backslash$
 \\
\hline
\end{tabular}
\end{table}

\subsection{Error Feedback Loop}
Due to the inherent hallucination tendencies of LLMs, even when comprehensive restrictions are provided to mitigate trivial errors, the same issues may persist. 
Inspired by real-world coding practices—where code is rarely correct on the first attempt, and iterative feedback from simulation and synthesis tools is critical for meeting design specifications—we adopt a feedback-based method that efficiently leverages error information from each query and the simulator.

For each problem, the initial query follows the standard process introduced in \Cref{sec:code_gen_eval}.
However, if the simulator detects a syntax error, our correction feedback loop is triggered. 
The error is first classified into specific categories, as outlined in \Cref{sec:Error_class}, enabling the precise identification of its cause without requiring the LLM to interpret abstract error messages which can directly inform code refinement. 
Next, the error category, along with detailed error reports and a crafted feedback prompt, is fed back to the LLM to refine the previously generated code in a manner similar to human debugging.
If the evaluation instead identifies a functional error, the feedback loop provides a concise prompt 
``The syntax is correct, but a functional error has occurred. Please review the problem description carefully". 
The feedback loop will continue iteratively until the code passes or the maximum number of iterations is reached.

\begin{figure}[!tb]
    \centering
    \subfloat{%
          \includegraphics[width=0.45\textwidth]{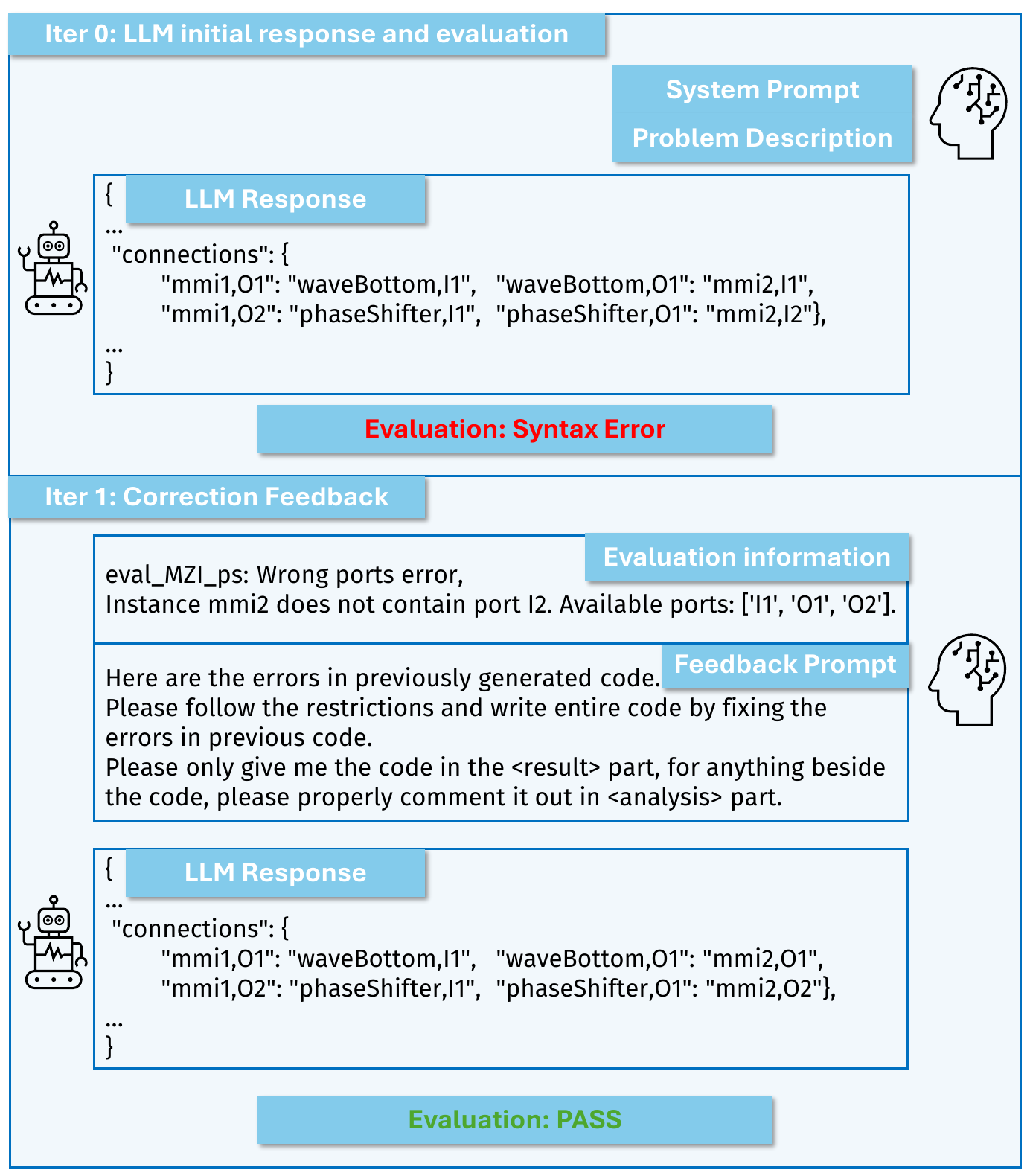}
      }
      \caption{An example of solving MZI\_ps by GPT o1-mini with feedback}
      \label{fig:feedback}
    \end{figure}

\Cref{fig:feedback} gives an example of MZI\_ps to illustrate this process,
Initially, the LLM generates a result that incorrectly connects to non-existent port I2. 
This error is automatically classified as a ``Wrong Ports Error." The classification, combined with detailed error information and a crafted feedback prompt, is then fed back to the LLM. 
After one iteration of the correction feedback loop, the error is resolved, and the code passes the test.

\section{Experimental Results}
\label{sec:results}
\subsection{Experiment Setup}
We implemented PICBench in Python.
The quality of the given PIC design was evaluated using the open-source simulation tool SAX, which specializes in S-parameter-based circuit simulations.
We constructed the S-parameters for essential devices, including waveguides, couplers, MMIs, MZIs, MRRs, and phase shifters, to simulate the frequency-domain response of the specified PIC over the wavelength range of 1510 to 1590 nm.

PICBench is compatible with a wide range of LLMs as long as they provide a Python API. 
In our experiment, we extensively evaluated the capabilities of five notable LLMs developed by leading companies, using PICBench: 
\begin{itemize}
    \item \textbf{GPT-4}: The free commercial model developed by OpenAI.
    \item \textbf{GPT-4o}: The flagship commercial model that is widely recognized for its versatility and high accuracy across various domains.
    \item \textbf{GPT-o1-mini}: A smaller GPT-based model optimized for STEM reasoning, with a focus on mathematical and technical problem-solving.
    \item \textbf{Gemini 1.5 Pro}: Developed by Google, an emerging model that integrates hybrid architectures to improve code generation and context handling.
    \item \textbf{Claude 3.5 Sonnet}: Developed by Anthropic, emphasizing significant improvement in graduate-level reasoning, knowledge acquisition, and coding abilities.
\end{itemize}

We adopted the widely used Pass@$k$ metric \cite{chen2021evaluating} to measure code generation correctness.
A problem is considered solved if any of the $k$-generated samples passes the corresponding unit tests.
For each task, n samples (default $n=5$) are generated, of which c samples pass, and an unbiased estimator of Pass@k is computed as:
    \begin{equation}
        \text{pass@k} := \mathbb{E}_{\text{Problems}} \left[ 1 - \frac{\binom{n-c}{k}}{\binom{n}{k}} \right].
    \label{eq:passk}
    \end{equation}
    
To analyze the impact of error feedback (EF), we queried the selected LLMs both without and with feedback for $n$ iterations where we set $n=1$ and $n=3$.   

\subsection{Results}

\begin{table*}[h!]
\caption{The Syntax and Functionality (Func.) evaluation for different LLMs. EF denotes the error feedback.}
\centering
\resizebox{\textwidth}{!}{
\begin{tabular}{ccccccccccccc}
\toprule
\multirow{4}{*}{\textbf{LLM}} & \multicolumn{6}{c}{\textbf{Pass@1}} & \multicolumn{6}{c}{\textbf{Pass@5}} \\ 
\cmidrule(lr){2-7} \cmidrule(lr){8-13}
& \multicolumn{2}{c}{\textbf{Without EF}} & \multicolumn{2}{c}{\textbf{With 1 EFs}} & \multicolumn{2}{c}{\textbf{With 3 EFs}} 
& \multicolumn{2}{c}{\textbf{Without EF}} & \multicolumn{2}{c}{\textbf{With 1 EFs}} & \multicolumn{2}{c}{\textbf{With 3 EFs}} \\
\cmidrule(lr){2-3} \cmidrule(lr){4-5} \cmidrule(lr){6-7} \cmidrule(lr){8-9} \cmidrule(lr){10-11} \cmidrule(lr){12-13}
& Syntax & Func. & Syntax & Func. & Syntax & Func. 
& Syntax & Func. & Syntax & Func. & Syntax & Func. \\
\midrule

GPT-4 & \bf{16.67} & 6.67 & 34.17 & 6.67 & 54.17 & 10.83 & \bf{41.67} & 12.50 & \bf{70.83} & 16.67 & \bf{100.00} & 29.17 \\
GPT-o1-mini & 8.33 & 4.17 & 33.33 & 15.00 & 63.33 & 23.33 & 29.17 & \bf{16.67} & 66.67 & \bf{25.00} & 91.67 & 33.33 \\
GPT-4o & 14.17 & 4.17 & \bf{40.83} & 15.00 & 59.17 & 20.00 & 37.50 & 4.17 & \bf{70.83} & \bf{25.00} & 87.50 & \bf{41.67} \\
Claude 3.5 Sonnet & 13.33 & 1.67 & 35.83 & 14.17 & \bf{75.83} & \bf{24.17} & 20.83 & 8.33 & \bf{70.83} & 20.83 & \bf{100.00} & 37.50 \\
Gemini 1.5 pro & 9.17 & \bf{8.33} & 33.33 & \bf{16.67} & 50.00 & 20.83 & 16.67 & 12.50 & 66.67 & 20.83 & 87.50 & 33.33 \\

\bottomrule
\end{tabular}}
\label{tab:wo_restrictions}
\end{table*}

\begin{table*}[h!]
\caption{The Syntax and Functionality (Func.) correctness evaluation for different LLMs with our proposed restrictions. EF denotes the error feedback.}
\centering
\resizebox{\textwidth}{!}{
\begin{tabular}{ccccccccccccc}
\toprule
\multirow{4}{*}{\textbf{LLM}} & \multicolumn{6}{c}{\textbf{Pass@1}} & \multicolumn{6}{c}{\textbf{Pass@5}} \\ 
\cmidrule(lr){2-7} \cmidrule(lr){8-13}
& \multicolumn{2}{c}{\textbf{Without EF}} & \multicolumn{2}{c}{\textbf{With 1 EFs}} & \multicolumn{2}{c}{\textbf{With 3 EFs}} 
& \multicolumn{2}{c}{\textbf{Without EF}} & \multicolumn{2}{c}{\textbf{With 1 EFs}} & \multicolumn{2}{c}{\textbf{With 3 EFs}} \\
\cmidrule(lr){2-3} \cmidrule(lr){4-5} \cmidrule(lr){6-7} \cmidrule(lr){8-9} \cmidrule(lr){10-11} \cmidrule(lr){12-13}
& Syntax & Func. & Syntax & Func. & Syntax & Func. 
& Syntax & Func. & Syntax & Func. & Syntax & Func. \\
\midrule

GPT-4 + restrictions & 20.00 & 4.17 & 38.33 & 17.50 & 71.67 & 30.00 & 58.33 & 12.50 & 58.33 & 20.83 & \bf{100.00} & 45.83 \\
GPT-o1-mini + restrictions & 13.33 & 9.17 & 50.00 & 22.50 & 84.17 & 33.33 & 25.00 & 12.50 & 79.17 & 33.33 & \bf{100.00} & 50.00 \\
GPT-4o + restrictions & 60.83 & 20.00 & 86.67 & 31.67 & \bf{95.00} & 36.67 & 87.50 & \bf{37.50} & \bf{100.00} & 37.50 & \bf{100.00} & 50.00 \\
Claude 3.5 Sonnet + restrictions & 54.17 & 15.83 & 80.83 & 29.00 & 90.00 & \bf{43.33} & 87.50 & \bf{37.50} & \bf{100.00} & \bf{45.83} & \bf{100.00} & \bf{62.50} \\
Gemini 1.5 pro + restrictions & \bf{64.17} & \bf{21.67} & \bf{88.33} & \bf{32.50} & \bf{95.00} & 38.33 & \bf{91.67} & \bf{37.50} & \bf{100.00} & \bf{45.83} & \bf{100.00} & 54.17 \\

\bottomrule
\end{tabular}}
\label{tab:w_restrictions}
\end{table*}

\subsubsection{Impact of error feedback}
\label{sec:error_feedback}

\Cref{tab:wo_restrictions} presents the comprehensive evaluation results for both syntax and functionality across all five selected LLMs, using 0, 1, and 3 error feedback iterations with PICBench, evaluated through the Pass@$k$ metric.

Without correction feedback, GPT-4 demonstrates the highest syntax accuracy, with 16.67\% for Pass@1 and 41.67\% for Pass@5, establishing its strength in pattern recognition and abstraction capabilities.

When error feedback is incorporated, there is a clear improvement in both syntax and functionality scores across Pass@1 and Pass@5 evaluations.
Claude 3.5 Sonnet demonstrated the most significant improvement in both syntax performance and functionality in response to feedback, highlighting its excellent self-correction and adaptive learning capabilities. 
At Pass@1, its syntax score increased from 13.33\% without feedback to 35.83\% with one feedback iteration, and further soared to 75.83\% with three iterations. 
Similarly, its functionality success rate improved from 1.67\% to 24.17\%.

Moreover, even one round of feedback can raise Pass@1 metrics beyond the Pass@5 results obtained without any feedback.
This further emphasizes the impact of feedback on model performance. 
For example, Gemini 1.5 pro achieves a syntax score of 33.33\% with one feedback iteration at Pass@1, which is notably higher than its Pass@5 syntax score without feedback (16.67\%).

Overall, the feedback method, aided by simulator diagnostics, rapidly accelerates debugging and enables iterative improvements in code generation. 
By revealing each model’s evolving logic and capacity for self-refinement, this approach delivers notable gains in both code quality and reliability.

\subsubsection{Impact of restrictions}
To investigate the impact of the restrictions introduced in \Cref{sec:Error_class}, we queried the selected LLMs using the same approach described in \Cref{sec:error_feedback}.

As shown in the \Cref{tab:w_restrictions}, the application of restrictions greatly enhances the performance of LLMs across all evaluated conditions, with notable improvements in both syntax and functionality. 
While all models benefit, the degree of improvement varies, with Gemini 1.5 pro showing the most dramatic enhancements across both syntax and functionality dimensions. 
For instance, Gemini 1.5 pro’s syntax score improves from 9.17\% to 64.17\% in Pass@1 and from 16.67\% to 91.67\% in Pass@5, while its functionality score increases from 8.33\% to 21.67\% in Pass@1 and from 12.50\% to 37.50\% in Pass@5 with restrictions applied, demonstrating its robust in-context reasoning and real-time adaptability.
The results highlight the potential for further optimization by leveraging the in-context learning ability of LLMs, enhanced with high-definition circuit knowledge.

\section{Conclusion}
\label{sec:conclu}
In this paper, we propose a comprehensive open-source benchmark for PIC design generation using LLMs, featuring 24 meticulously crafted PIC design problems. 
We also introduce a feedback-based prompt engineering technique that iteratively refines designs and enhances the models' design generation capabilities.
Our comprehensive evaluation of various state-of-the-art commercial LLMs highlights the significant impact of feedback mechanisms and the utilization of in-context learning capabilities on model performance. 
The results underscore both the challenges and opportunities for LLMs in the PIC design domain, providing targeted insights for future research and development to advance automation and efficiency in this field.

\section*{Acknowledgment}
\label{sec:ack}

This work is supported by the Nansha District Key Area S\&T Scheme (No. 2024ZD007), Guangzhou Municipal Science and Technology Project (Guangzhou EDA Key Laboratory, No.2023A03J0013), and Natural Science Foundation of Guangdong Province (No.2024A1515012438).
%\printbibliography
% \newpage
% {
%\bibliographystyle{ACM-Reference-Format}
%\bibliographystyle{splncs04}
% \bibliographystyle{IEEEtran}
% \bibliography{ref/Top-sim,ref/PDA}

% \bibliographystyle{IEEEtran}
% \bibliography{ref/Top-sim,ref/PDA}
% }
\end{document}